\documentclass[acmtog,nonacm]{acmart}

\acmSubmissionID{1019}
\acmJournal{TOG}

\usepackage{booktabs} 
\usepackage{stfloats}
\usepackage[ruled]{algorithm2e} 

\SetAlFnt{\small}
\SetAlCapFnt{\small}
\SetAlCapNameFnt{\small}
\SetAlCapHSkip{0pt}
\usepackage{xspace}
\usepackage{multirow}
\usepackage{adjustbox}

\makeatletter
\DeclareRobustCommand\onedot{\futurelet\@let@token\@onedot}
\def\@onedot{\ifx\@let@token.\else.\null\fi\xspace}

\def\eg{\emph{e.g}\onedot} 
\def\ie{\emph{i.e}\onedot}

\makeatother

\begin{document}

\title{InstructAV2AV: Instruction-Guided Audio-Video Joint Editing}

\author{Haojie Zheng}
\affiliation{%
  \institution{Beijing Academy of Artificial Intelligence}
  \city{Beijing}
  \country{China}
}
\affiliation{%
  \institution{Peking University}
  \city{Beijing}
  \country{China}
}

\email{suimu@stu.pku.edu.cn}

\author{Yixin Yang}
\affiliation{%
  \institution{Peking University}
  \city{Beijing}
  \country{China}
}
\email{yangyixin93@pku.edu.cn}

\author{Siqi Yang}
\affiliation{%
  \institution{Peking University}
  \city{Beijing}
  \country{China}
}
\email{yousiki@pku.edu.cn}

\author{Shuchen Weng}
\authornote{Corresponding authors.}
\affiliation{%
  \institution{Beijing Academy of Artificial Intelligence}
  \city{Beijing}
  \country{China}
}
\affiliation{%
  \institution{Peking University}
  \city{Beijing}
  \country{China}
}
\email{shuchenweng@pku.edu.cn}

\author{Boxin Shi}
\authornotemark[1]
\affiliation{%
  \institution{Peking University}
  \city{Beijing}
  \country{China}
}
\email{shiboxin@pku.edu.cn}

\renewcommand\shortauthors{Haojie Zheng, Yixin Yang, Siqi Yang, Shuchen Weng, and Boxin Shi}

\begin{teaserfigure}
  \centering
  \includegraphics[width=\linewidth]{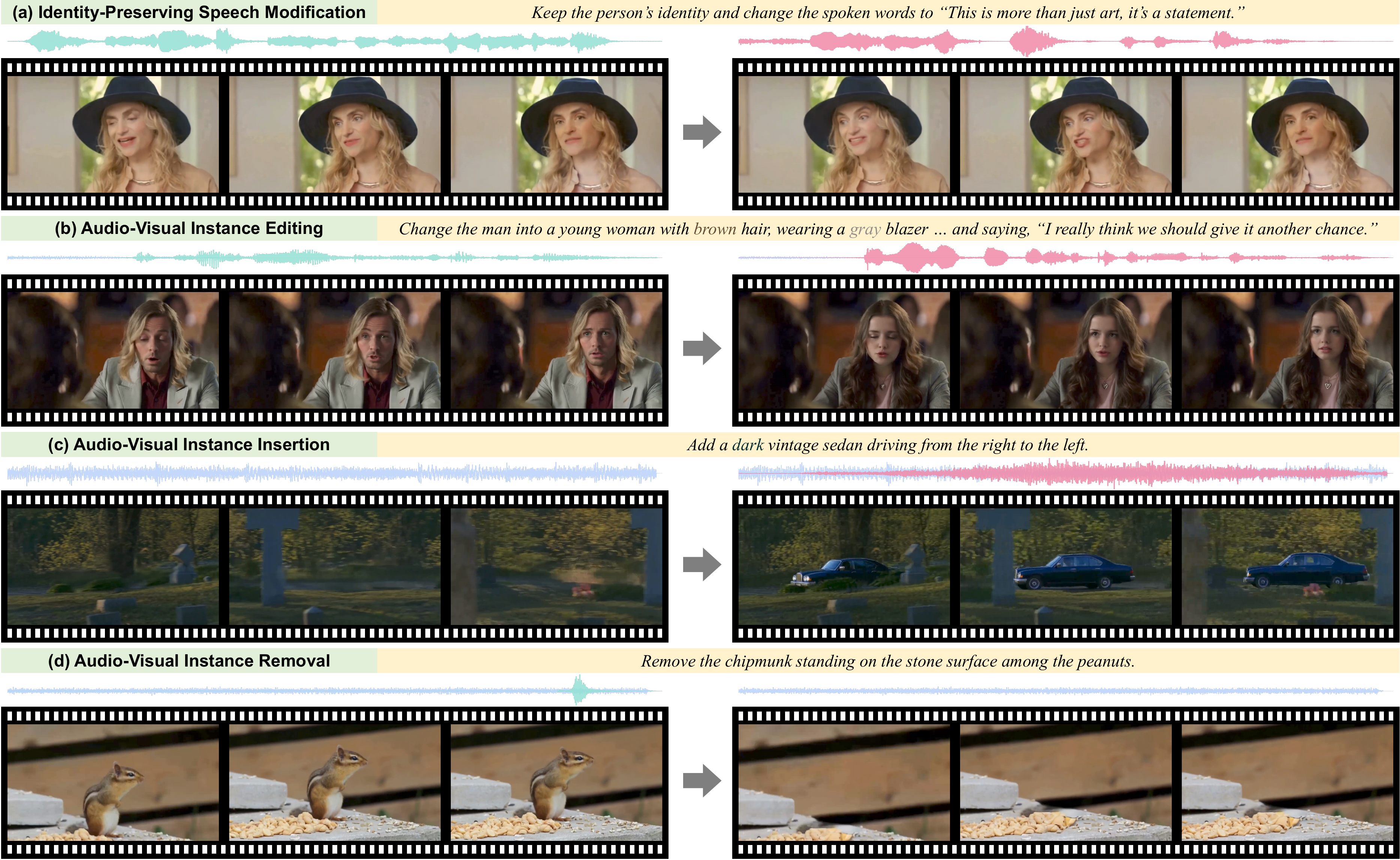}
  \vspace{-7mm}
  \caption{
  InstructAV2AV achieves open-world instruction-guided audio-video joint editing. Given solely text instructions, it manipulates specified visual targets and their audio tracks while faithfully preserving non-target backgrounds and ambient sounds. 
  We demonstrate its versatility across four representative scenarios: (a) Identity-preserving speech modification, changing the spoken words of the woman to a target sentence while strictly retaining her visual identity and vocal timbre; (b) Audio-visual instance editing, replacing the speaking man with a young woman in a gray blazer alongside her synchronized voice; (c) Audio-visual instance insertion, adding a dark vintage sedan driving across the scene and its engine sound; and (d) Audio-visual instance removal, erasing the chipmunk and its scream from the environment. 
  We highly encourage readers to view the supplementary video for dynamic results.
  }
  \label{fig:teaser}
\end{teaserfigure}

\begin{abstract}
Recent diffusion-based methods have achieved impressive progress in video content manipulation. However, they typically ignore the accompanying audio, leaving the audio disjointed from the edited results. 
In this paper, we propose InstructAV2AV, the first end-to-end framework for instruction-guided audio-video joint editing. 
We first develop a scalable data synthesis pipeline and construct InsAVE-80K, the first large-scale audio-video editing dataset with high-quality source-to-target pairs. 
With this data foundation, we adapt an audio-video generation backbone to leverage its robust priors. We concatenate the audio-video input with noisy latent codes to anchor the source context, propose the source-instruction gated attention to improve instruction following and content preservation, and introduce a two-stage training strategy to effectively transfer these pre-trained priors. 
Extensive experiments demonstrate that InstructAV2AV outperforms state-of-the-art methods across 11 metrics spanning three aspects on two evaluation sets, highlighting its potential for controllable content creation.
Project page: \href{https://hjzheng.net/projects/InstructAV2AV/}{\textcolor{teal}{https://hjzheng.net/projects/InstructAV2AV/}}
\end{abstract}



\maketitle

\section{Introduction}
Recent advances in diffusion-based video editing~\cite{senorita, insvie, ditto} have enabled complex visual content manipulations guided by text instructions, demonstrating immense potential in practical applications (\eg, content creation~\cite{dreamix}, visual effects~\cite{magicvfx}, and environment simulation~\cite{drivedreamer}). However, real-world media is inherently multimodal, where visual content is tightly synchronized with its accompanying audio. Existing video editing methods primarily manipulate the visual modality while leaving the audio unchanged. Consequently, when the edited visual content alters the identity, action, or event source in a video, the original audio becomes temporally or semantically disjointed from the edited result. This limitation fundamentally motivates the exploration of audio-video editing methods~\cite{aviedit, coherentavedit}, which aim to concurrently modify user-specified targets across both modalities to produce coherent audio-visual results.

However, existing audio-video editing methods still struggle with fine-grained controllability.
Many prior efforts~\cite{liang2024language} primarily focus on global scene transformations, which fundamentally restricts their ability to selectively manipulate a specific object and its accompanying audio.
While recent methods~\cite{aved,aviedit} achieve instance awareness through the introduction of spatial masks, this introduces tedious manual annotation costs and hinders interactive applications.
Other paradigms are either confined to specific operations (\eg, object removal~\cite{save}) or heavily reliant on dense auxiliary conditions (\eg, Canny edges~\cite{avcontrol}).
These limitations highlight the urgent need for an instruction-guided framework capable of performing open-world audio-video joint editing using solely text instructions, while inherently preserving non-target contexts and maintaining strict audio-video synchronization.

In this paper, we propose \textbf{InstructAV2AV}, an end-to-end framework for instruction-guided audio-video joint editing.
As illustrated in Figure~\ref{fig:teaser}, given a source audio-video sample and a text instruction as inputs, InstructAV2AV jointly manipulates the specified visual target and its corresponding audio track, while faithfully preserving the original background, irrelevant objects, and non-target ambient sounds.
This paradigm enables a wide range of open-world audio-video editing applications, including modifying the speech of an active speaker (Figure~\ref{fig:teaser}~(a)), jointly replacing a visual object and its sound source (Figure~\ref{fig:teaser}~(b)), inserting a new object and its synchronized audio (Figure~\ref{fig:teaser}~(c)), and removing a specified instance and its acoustic footprint (Figure~\ref{fig:teaser}~(d)).

Building a solid foundation for InstructAV2AV requires extensive training data. 
Given the scarcity of such source-to-target paired resources in the audio-visual domain, we design a scalable data synthesis pipeline. 
Specifically, we develop a mask-guided model for audio-sync video editing as our data engine. Following the strategy of AVI-Edit~\cite{aviedit}, we optimize this model using source data, thereby generating large-scale paired target samples. 
To ensure the fidelity of the synthesized data, we further introduce a comprehensive filtering process with five rigorous criteria to discard suboptimal samples, alongside additional human verification to enable a highly reliable benchmark. 
Ultimately, this pipeline produces \textbf{InsAVE-80K}, the first large-scale instruction-guided audio-video editing dataset, comprising 79K training and 1K evaluation samples that include source media, synthesized targets, and text instructions.

Leveraging this foundational dataset, we build InstructAV2AV upon a pre-trained audio-video generation model~\cite{ovi} and adapt it for instruction-guided joint editing. By conditioning on the source audio and video latent codes as contextual references, InstructAV2AV learns to predict the edited targets under the guidance of text instructions.
We propose the Source-Instruction Gated Attention (SIGA) module to improve content preservation and instruction following via a learnable soft gate, enabling the noisy latent code to concurrently integrate source features and the text instruction.
We further propose a two-stage training strategy to effectively transfer the generative priors to the editing domain. By independently adapting the video and audio branches before conducting joint optimization, we facilitate smooth and robust model convergence.
Extensive experiments demonstrate the effectiveness of InstructAV2AV for open-world audio-video joint editing.

Our contributions are summarized as follows:

\begin{itemize}
    \item We propose InstructAV2AV, an end-to-end instruction-guided audio-video joint editing framework, enabling open-world fine-grained manipulations using solely text instructions.
    
    \item We construct InsAVE-80K, the first large-scale audio-video editing dataset with source-to-target pairs and quality filtering, as a data foundation for model training and evaluation.
    
    \item We propose the SIGA module to improve instruction following and content preservation via a soft gate, and the two-stage training strategy for smooth model convergence.

\end{itemize}

\section{Related Work}

\subsection{Instruction-guided Video Editing}
The instruction-guided paradigm~\cite{instructpix2pix,prompt2prompt,imagic} originates in the image domain (\eg, InstructPix2Pix~\cite{instructpix2pix}), where models edit visual content based solely on text instructions rather than dense structural conditions, thereby significantly reducing the user burden.
With the rapid development of large-scale video generation models~\cite{wan, hunyuan-video, cogvideox}, this editing paradigm extends into the video field. 
Early video editing works rely on test-time optimization techniques~\cite{CoDeF, tune-a-video}, which achieve impressive temporal consistency but incur prohibitively high computational costs during inference. 
To address this inefficiency, researchers explore feed-forward InstructVid2Vid~\cite{instructvid2vid}. However, these early attempts rely on simplistic frame-level data synthesis, struggling with limited dataset diversity. 
To overcome the data bottleneck, more recent works~\cite{senorita, insvie, ditto} introduce scalable data generation pipelines. For instance, Señorita-2M~\cite{senorita} employs a complex pipeline cascading multiple specialized vision expert models to construct high-quality editing pairs.
Despite these remarkable improvements, existing methods strictly focus on the visual modality. Therefore, we propose InstructAV2AV to address the neglect of acoustic elements in real-world media, ensuring that the edited visuals maintain temporally and semantically alignment with their accompanying audio.

\subsection{Audio-Video Joint Generation}
Pioneering audio-video joint generation models~\cite{mmdiffusion,liu2023sounding} propose unified frameworks to synthesize audio and video simultaneously. 
To pursue highly realistic and immersive media, tech companies have developed powerful commercial models~\cite{sora2,veo3} capable of generating high-fidelity audio-video results. 
These commercial successes further motivate the research community to advance open-source foundations. 
For instance, Ovi~\cite{ovi} adopts a symmetric twin-backbone diffusion transformer architecture, while UniVerse-1~\cite{universe-1} and MOVA~\cite{mova} leverage mixture-of-experts strategies to effectively fuse pretrained video and audio representations. Furthermore, recent advancements like LTX-2~\cite{ltx2} and the JavisDiT series~\cite{javisdit, liu2026javisdit++} significantly enhance cross-modal alignment through decoupled dual-stream architectures and hierarchical spatio-temporal prior synchronization. 
Despite these remarkable generative capabilities, directly transferring the generative priors of these foundational models to editing tasks remains highly challenging.
Therefore, we design a scalable data synthesis pipeline to serve as a solid data foundation, enabling us to effectively develop the first instruction-guided audio-video joint editing framework.

\subsection{Audio-Video Joint Editing}
Audio-video joint editing fundamentally requires synchronizing cross-modal semantic changes while strictly preserving non-target inherent content. 
Early methods~\cite{liang2024language} attempt to align audio-visual features by designing explicit cross-modal attention mechanisms for global scene transitions rather than instance-level control. 
To achieve localized manipulations, CoherentAVEditing~\cite{coherentavedit} constructs a multi-stage pipeline to sequentially generate edited visuals and subsequently synthesize the corresponding audio track. 
Other paradigms~\cite{aved, avcontrol, aviedit} modify the generation backbone for explicit spatial constraints. For instance, AVI-Edit~\cite{aviedit} injects user-provided spatial masks into the generation process, while AVControl~\cite{avcontrol} optimizes separately trained LoRA adapters for conditional structural alignment. 
Unlike these existing paradigms that rely on global scene transformations, cascaded generation pipelines, and dense structural injections, InstructAV2AV introduces a unified architectural solution. By integrating the proposed Source-Instruction Gated Attention (SIGA) module and a tailored two-stage joint training strategy, our framework achieves open-world fine-grained audio-video manipulations guided solely by text instructions in an end-to-end manner.

\begin{figure*}[t]
  \centering
  \includegraphics[width=\linewidth]{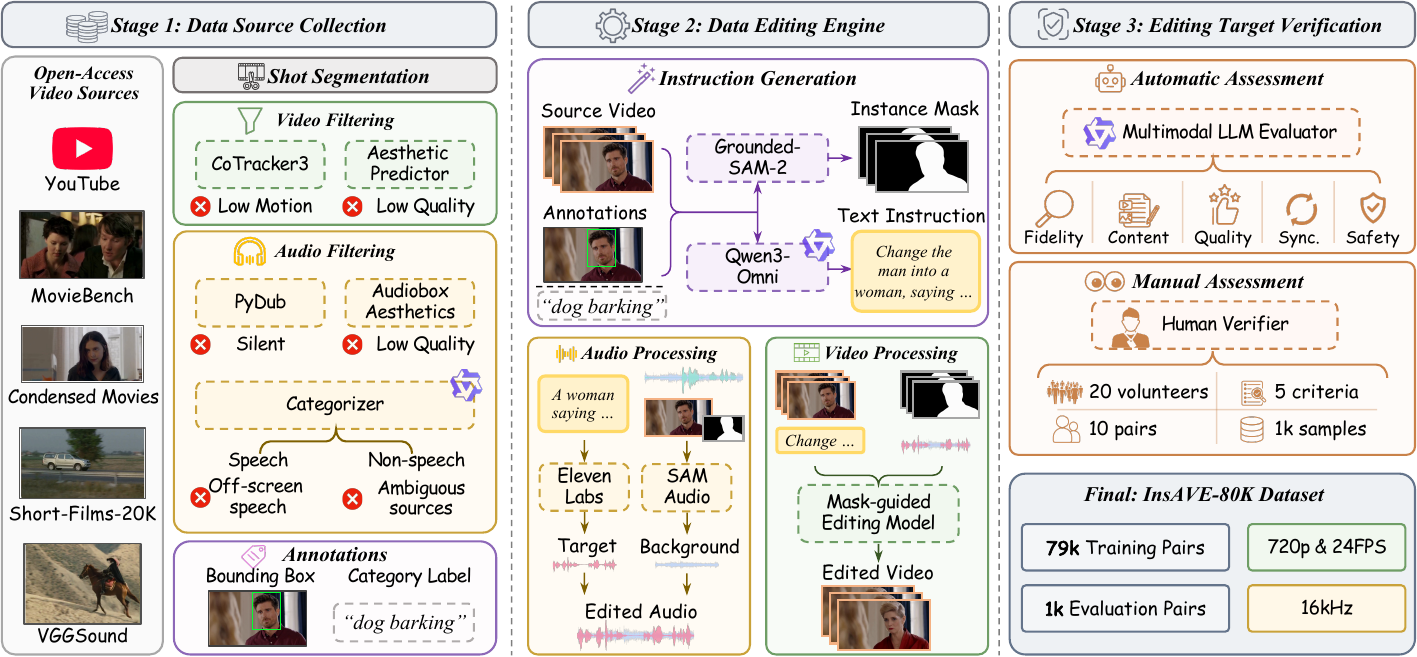}
  \vspace{-6pt}
  \caption{Overview of the scalable data synthesis pipeline for InsAVE-80K. (a) We curate open-world videos through a multi-stage audio-visual filtering process to ensure high-fidelity video quality and unambiguous audio sources. (b) We annotate instance masks and formulate text instructions, which guide the large-scale synthesis of target audio and corresponding videos, while preserving non-target components. (c) After applying a rigorous multimodal LLM assessment and additional human verification across five criteria, we ultimately compile the highly reliable InsAVE-80K dataset.}
  \vspace{-6pt}
  \label{fig:dataset_pipeline}
\end{figure*}

\section{Scalable Data Synthesis Pipeline}

Given the scarcity of high-quality paired editing data, training instruction-guided audio-video joint editing models at scale remains a formidable challenge. To overcome this bottleneck, we propose a scalable data synthesis pipeline to transform open-world raw media into highly verified source-target pairs. 

As illustrated in Figure~\ref{fig:dataset_pipeline}, our pipeline includes three progressive stages.
First, we curate a diverse collection of source clips, applying multi-stage filtering to ensure high-fidelity data quality and broad semantic coverage (Sec.~\ref{subsec:collection}). 
Second, we introduce a mask-guided data editing engine that automatically synthesizes editing instructions and their corresponding edited targets, generating highly plausible target data at scale (Sec.~\ref{subsec:engine}). 
Finally, we apply a comprehensive verification process to these synthesized samples, combining automatic quality evaluation with targeted manual inspection (Sec.~\ref{subsec:verification}). 
We compile these verified samples to construct InsAVE-80K, the first large-scale foundational dataset tailored for instruction-guided joint audio-video editing models.

\subsection{Data Source Collection} \label{subsec:collection}
We collect audio-video samples from publicly accessible online platforms (\eg, YouTube) and publicly available datasets, including MovieBench~\cite{moviebench}, Condensed Movies~\cite{cdm}, Short-Films-20K~\cite{sf20k}, and VGGSound~\cite{vggsound}. To improve data quality and ensure reliable audio-visual correspondence, we apply a multi-stage preprocessing pipeline.

\paragraph{Video filtering.}
Following previous works~\cite{svd, mtv, MIDAS}, we first use PySceneDetect~\cite{pyscenedetect} to segment raw videos into single-shot clips, effectively reducing abrupt shot transitions within each sample. Next, we adopt a tracking-based approach to estimate frame-level motion. Specifically, we sample points on a grid layout and track them using CoTracker3~\cite{cotracker3} to obtain point trajectories. We then compute the average frame-to-frame motion magnitude across all tracked points over the entire video. Clips with an average motion magnitude below a predefined threshold are discarded as static or low-motion samples. Finally, we use LAION Aesthetics Predictor~\cite{laion-aesthetic-predictor} to filter visually low-quality clips.

\paragraph{Audio filtering.}
We begin by removing silent clips below -45 dBFS using PyDub~\cite{pydub} and discard low-quality audio via Audiobox-Aesthetics~\cite{audiobox_aesthetics}. We then utilize Qwen3-Omni~\cite{qwen_omni} to categorize the remaining clips for content-specific filtering: \textit{(i)} For speech-dominated clips, we leverage TalkNet~\cite{talknet} to localize active speakers and Scribe~\cite{scribe} to extract precise speech timestamps. To reduce off-screen speech, clips are retained only when the speech is temporally aligned with a visible on-screen speaker. \textit{(ii)} For non-speech event clips, we filter out ambiguous sound sources to preserve semantic clarity. Ultimately, each retained clip is assigned a distinct semantic sound-event label (\eg, ``dog barking'') based on its dominant sound source.

\subsection{Data Editing Engine} \label{subsec:data_engine_1}

We design a data engine\footnote{Please refer to supplementary materials for more details.} to automatically transform open-world clips into high-quality training data at scale. Given a source clip, it localizes candidate sound sources, formulates fine-grained text instructions, and generates the corresponding audio-visual targets while strictly preserving non-target content.

\paragraph{Instruction generation.}
For each source clip, we first use Grounded-SAM-2~\cite{grounded_sam2} to obtain instance-level masks for visual entities, providing spatial grounding for subsequent manipulations. Conditioned on these candidates and the source audio-visual context, we employ Qwen3-Omni~\cite{qwen_omni} to formulate diverse text instructions for comprehensive editing operations.

\paragraph{Audio processing.}
Given the entity mask, we use SAM-Audio~\cite{sam_audio} to separate its original sound from the background audio. Guided by the generated instruction, a text-to-audio model~\cite{elevenlabs} is used to synthesize the corresponding edited soundtrack. By seamlessly mixing this synthesized track with the background audio, we obtain the target audio track that faithfully preserves non-target acoustic contexts.

\paragraph{Video processing.}
We further develop a mask-guided editing model to synthesize audio-synchronized target videos based on the generated instruction. Built upon Wan2.2-5B~\cite{wan2.2}, the model utilizes the entity mask to restrict spatial modifications.
We incorporate audio features into the video backbone via frame-wise cross-attention, ensuring strict temporal synchronization. Following AVI-Edit~\cite{aviedit}, we use a source-conditioned flow matching objective, thereby eliminating manual annotation for the engine's model optimization.

\subsection{Editing Target Verification} \label{subsec:verification}
To prevent the synthesized editing targets from suffering from instruction mismatch, perceptual artifacts, or unintended contextual changes, we introduce a rigorous verification pipeline to filter candidate source-target pairs and construct a highly reliable dataset.

\paragraph{Automatic assessment.}
We utilize a multimodal LLM~\cite{qwen_omni} to score each generated audio-video target along five dimensions: \textit{(i)} instruction fidelity for strict prompt adherence; \textit{(ii)} content preservation for unaltered non-target elements; \textit{(iii)} perceptual quality for realism; \textit{(iv)} audio-video synchronization for cross-modal alignment; and \textit{(v)} safety to filter inappropriate content. We retain only the samples that simultaneously pass all five criteria.

\paragraph{Manual assessment.}
To construct the evaluation set, we perform additional human verification on the automatically retained samples. Specifically, 20 volunteers are divided into 10 judge pairs. Each candidate sample is evaluated by five independent pairs, with one pair assigned to each criterion. A criterion is passed only when both judges in the pair agree. A sample is included in the evaluation set only if it passes all five criteria. This process is repeated until 1K high-quality evaluation samples are collected.

\paragraph{Dataset summary.}
We compile the verified source clips, editing instructions, and edited audio-video targets into InsAVE-80K, the \textbf{Ins}truction-guided \textbf{A}udio-\textbf{V}ideo joint \textbf{E}diting dataset. 
InsAVE-80K comprises 79K auto-verified training and 1K human-curated evaluation pairs, containing 5s clips at 720p/24FPS with 16kHz audio. To our best knowledge, it is the first large-scale dataset with source-target pairs for instruction-guided audio-video joint editing.

\begin{figure*}[t]
  \centering
  \includegraphics[width=\linewidth]{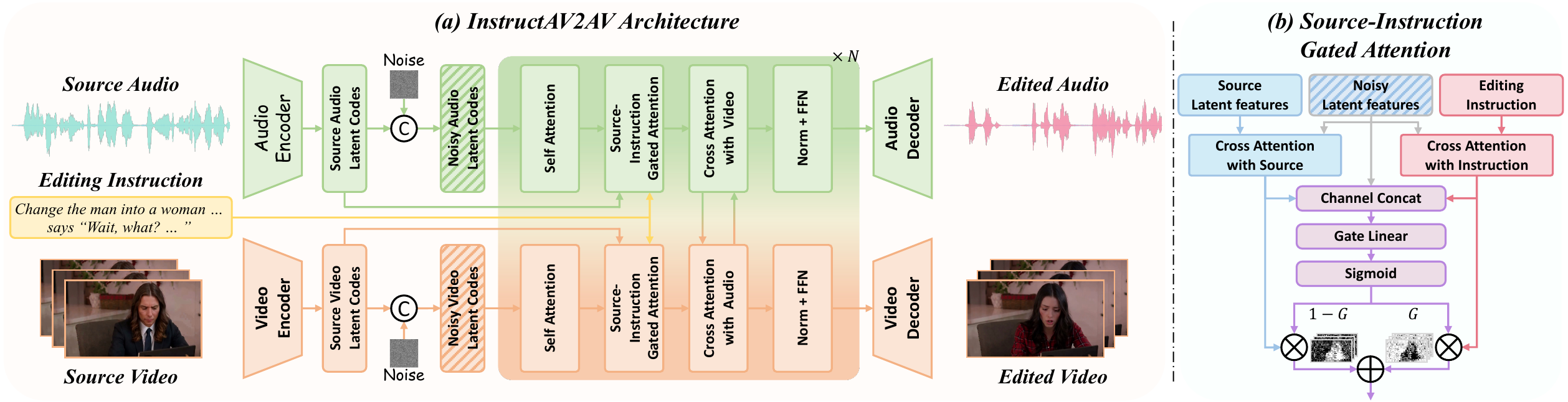}
  
  \caption{Illustration of our InstructAV2AV framework. Given audio-video sources and editing instructions, our framework effectively executes fine-grained manipulations while preserving non-target contexts. (a) InstructAV2AV adopts a pre-trained dual-stream diffusion transformer architecture. Within each block, the modalities first independently process internal dependencies via self-attention. The text instructions are then injected through our proposed SIGA module, followed by bidirectional cross-modal attention to ensure strict temporal and semantic synchronization. (b) In the SIGA module, the noisy latent concurrently interacts with both source and instruction features. The outputs are integrated via a learned soft gate, adaptively shifting the focus between content preservation and instruction following.}
  \label{fig:pipeline}
\end{figure*}

\section{InstructAV2AV}
\label{sec:instructav2av}

Leveraging the large-scale source-to-target pairs in InsAVE-80K as the data foundation, we propose \textbf{InstructAV2AV}, an end-to-end instruction-guided framework for open-world audio-video joint editing. 
First, we detail the approach to architecturally adapt an audio-video generation model into an editing framework (Sec.~\ref{sec:overview}).
Then, we introduce Source-Instruction Gated Attention (SIGA) to improve content preservation and instruction following via a learnable soft gate (Sec.~\ref{sec:siga}).
Finally, we present a two-stage training strategy to effectively transfer the generative priors to the editing domain while enabling smooth convergence (Sec.~\ref{sec:training_strategy}).

\subsection{Dual-Stream Editing Backbone}
\label{sec:overview}

We formulate instruction-guided audio-video joint editing as a conditional generation process. As illustrated in Figure~\ref{fig:pipeline}~(a), InstructAV2AV takes audio-video sources $\{x^\mathrm{v}, x^\mathrm{a}\}$ and text instructions $I$ as inputs to generate the edited audio-video results $\{y^\mathrm{v}, y^\mathrm{a}\}$. Initially, the source video and audio are encoded into latent codes $\{z_\mathrm{s}^\mathrm{v}, z_\mathrm{s}^\mathrm{a}\}=\{\mathcal{E}^\mathrm{v}(x^\mathrm{v}), \mathcal{E}^\mathrm{a}(x^\mathrm{a})\}$ using pre-trained spatial-temporal~\cite{wan} and 1D~\cite{mmaudio} VAE encoders, respectively.

\paragraph{Architecture adaptation.}
To strictly anchor the editing process to the source audio-video context, we concatenate the source latent codes and the Gaussian noise along the channel dimension for each modality. 
Structurally, InstructAV2AV adopts a symmetric dual-stream diffusion transformer architecture following the standard audio-video generation model~\cite{ovi}. 
Within this framework, each modality first independently applies self-attention to model its internal dependencies, capturing dynamic patterns. 
To guide the semantic direction for fine-grained manipulations, the editing instructions $I$ are encoded via T5~\cite{T5} and injected into the backbone through our proposed Source-Instruction Gated Attention (SIGA).
Finally, the video and audio branches exchange features via bidirectional cross-modal attention, ensuring joint modeling for strict temporal and semantic synchronization.

\paragraph{Flow matching formulation.}
We formulate our dual-stream diffusion transformer $v_{\theta}$ within a flow matching framework~\cite{flow_matching}.
Given Gaussian noise $\epsilon^\mathrm{v}, \epsilon^\mathrm{a} \sim \mathcal{N}(0,\mathbf{I})$, target edited latent codes $\{z_1^\mathrm{v},z_1^\mathrm{a}\}=\{\mathcal{E}^\mathrm{v}(y^\mathrm{v}),\mathcal{E}^\mathrm{a}(y^\mathrm{a})\}$, and a time step $t\in[0,1]$, the intermediate noisy latent codes are constructed as:
\begin{equation}
z_t^\mathrm{v}=(1-t)\epsilon^\mathrm{v}+t z_1^\mathrm{v}, \qquad z_t^\mathrm{a}=(1-t)\epsilon^\mathrm{a}+t z_1^\mathrm{a}.
\end{equation}
Conditioned on the source latent codes $\{z_\mathrm{s}^\mathrm{v}, z_\mathrm{s}^\mathrm{a}\}$ and the instruction embedding $C_\mathrm{I}$, we train the network to predict the target velocity fields $u_t^\mathrm{v}=z_1^\mathrm{v}-\epsilon^\mathrm{v}$ and $u_t^\mathrm{a}=z_1^\mathrm{a}-\epsilon^\mathrm{a}$:
\begin{equation}
(\hat{u}_t^\mathrm{v}, \hat{u}_t^\mathrm{a}) = v_{\theta}\left(t, (z_t^\mathrm{v}, z_t^\mathrm{a}), (z_\mathrm{s}^\mathrm{v}, z_\mathrm{s}^\mathrm{a}), C_\mathrm{I}\right).
\end{equation}
We then optimize the diffusion transformer by minimizing the loss:
\begin{equation}
\mathcal{L}_{\mathrm{FM}} = \mathbb{E}_{t, \{\epsilon^m,z_1^m,z_\mathrm{s}^m\}_{m \in \{\mathrm{v}, \mathrm{a}\}}, I} \left[ \lambda_\mathrm{v}\left\|\hat{u}_t^\mathrm{v}-u_t^\mathrm{v}\right\|_2^2 + \lambda_\mathrm{a}\left\|\hat{u}_t^\mathrm{a}-u_t^\mathrm{a}\right\|_2^2 \right],
\end{equation}
where $m \in \{\mathrm{v}, \mathrm{a}\}$ denotes the video and audio modalities, and $\lambda_\mathrm{v}=0.85, \lambda_\mathrm{a} = 0.15$ are the empirical modality-balancing weights.

\paragraph{Inference.}
We sample Gaussian noise $z_0^\mathrm{v}, z_0^\mathrm{a} \sim \mathcal{N}(0,\mathbf{I})$ during inference, and transform them into edited clean latent codes by solving:
\begin{equation}
\frac{\mathrm{d}}{\mathrm{d}t} (z_t^\mathrm{v}, z_t^\mathrm{a}) = v_\theta\left(t, (z_t^\mathrm{v}, z_t^\mathrm{a}), (z_\mathrm{s}^\mathrm{v}, z_\mathrm{s}^\mathrm{a}), C_\mathrm{I}\right),
\end{equation}
from $t=0$ to $t=1$ using a numerical solver.
The final denoised latent codes $\{z_1^\mathrm{v}, z_1^\mathrm{a}\}$ are then passed through their respective pre-trained VAE decoders to reconstruct the edited audio-video results $\{y^\mathrm{v}, y^\mathrm{a}\} = \{\mathcal{D}^\mathrm{v}(z_1^\mathrm{v}), \mathcal{D}^\mathrm{a}(z_1^\mathrm{a})\}$.

\subsection{Source-Instruction Gated Attention}
\label{sec:siga}

For the instruction-guided audio-video editing task, the model should preserve the original audio-video context while following the instruction context for the target semantic direction. However, since InstructAV2AV relies solely on text instructions, it lacks explicit spatial-temporal masks to localize the editing regions. Therefore, we propose the Source-Instruction Gated Attention (SIGA) to adaptively determine the dominant context for each individual token, where the noisy latent code serves as a query to interact with both source and instruction features via a learned soft gate.

As shown in Figure~\ref{fig:pipeline}~(b), let $f_\mathrm{h}$ denote the current noisy latent features in a DiT block, $f_\mathrm{x}$ denote the source latent features, and $f_\mathrm{c}$ denote the instruction features. To explicitly formulate the unified query mechanism, the noisy latent $f_\mathrm{h}$ is projected into a shared query $Q = f_\mathrm{h} W_{Q}$, while the conditions are projected into their respective key-value pairs: $(K_\mathrm{x}, V_\mathrm{x})$ from $f_\mathrm{x}$, and $(K_\mathrm{c}, V_\mathrm{c})$ from $f_\mathrm{c}$. With these projections, SIGA concurrently attends to both conditional streams:
\begin{equation}
    \hat{f}_\mathrm{h}^m = \operatorname{Attn}(Q, K_m, V_m), \quad \text{for} \quad m \in \{\mathrm{x}, \mathrm{c}\},
\end{equation}
where $\operatorname{Attn}(\cdot)$ denotes the standard scaled dot-product attention, $\hat{f}_\mathrm{h}^\mathrm{x}$ is responsible for preserving the original context, and $\hat{f}_\mathrm{h}^\mathrm{c}$ represents the targeted semantic modifications.
To adaptively integrate these two contexts, SIGA employs a learned soft gate to predict a token-wise gating map $G$ and fuses the outputs:
\begin{equation}
G = \sigma \left( W_g [ \hat{f}_\mathrm{h}^\mathrm{x}; \hat{f}_\mathrm{h}^\mathrm{c}; f_\mathrm{h} ] \right), \qquad \tilde{f}_\mathrm{h} = (1-G)\odot \hat{f}_\mathrm{h}^\mathrm{x} + G \odot \hat{f}_\mathrm{h}^\mathrm{c},
\end{equation}
where $[\cdot;\cdot]$ denotes channel-wise concatenation, $W_g$ is a learnable linear projection, $\sigma(\cdot)$ is the sigmoid activation, and $\odot$ denotes element-wise multiplication.
As $G$ approaches zero, the network prioritizes the source features to preserve non-target content. Conversely, as $G$ approaches one, it shifts focus to the instruction features to execute targeted modifications.

\subsection{Two-Stage Training Strategy}
\label{sec:training_strategy}

Directly fine-tuning the full audio-video model for joint editing presents convergence challenges. It requires the network to simultaneously learn source preservation, instruction following, and complex cross-modal synchronization, which complicates the optimization process. To effectively transfer the pre-trained generative priors to the editing domain, we present a two-stage training strategy.

\paragraph{Modality-specific training.}
In the first stage, we decouple the audio and video branches to train them independently. Specifically, we bypass the bidirectional cross attention between the two streams. During optimization, we apply the flow matching objective to each modality independently by updating one branch and omitting the other. This decoupled training forces each branch to focus on transferring its modality prior without cross-modal interference.

\paragraph{Joint audio-video training.}
In the second stage, we unlock the full architecture and jointly finetune the model with the bidirectional cross-attention enabled. Initialized with the weights from the first stage, the network learns to synchronize the visual and acoustic edits. This forces the model to capture the fine-grained correlations between visual movements and acoustic events, ensuring strict temporal and semantic coherence in the final results.

\begin{figure*}[t]
  \centering
  \includegraphics[width=\linewidth]{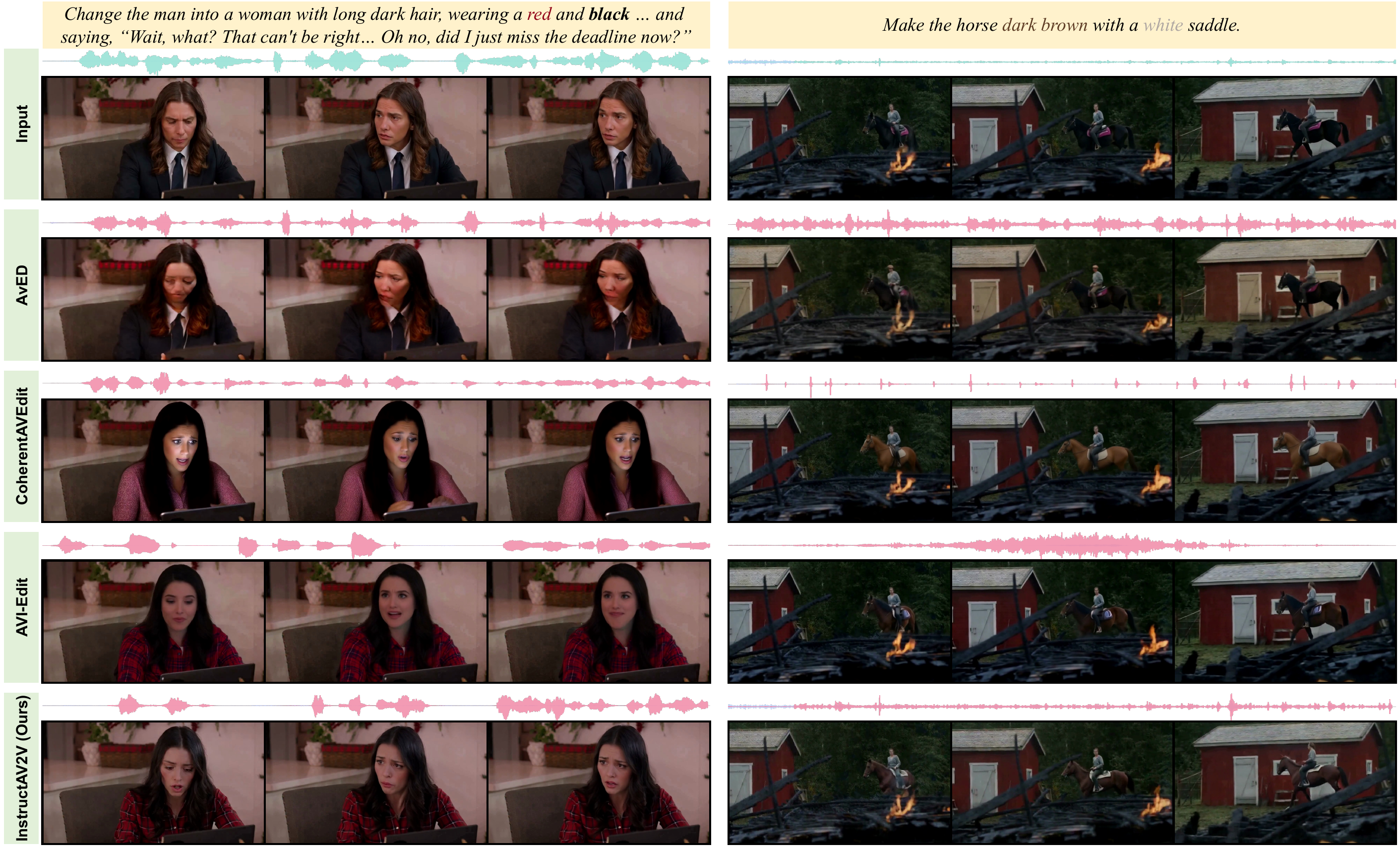}
  \caption{
   Qualitative comparison results with state-of-the-art methods for audio-video joint editing.
  }
  \label{fig:compare}
\end{figure*}

\section{Experiments}

\begin{table*}[t]
\caption{Quantitative experiment results of comparison and ablation. $\uparrow$ ($\downarrow$) means higher (lower) is better. Best performances are highlighted in \textbf{bold}.}
\label{tab:quantitative_combined}
  \vspace{-3pt}

\begin{center}
\small
\setlength\tabcolsep{2pt}
\begin{adjustbox}{width=\textwidth,keepaspectratio}

\begin{tabular}{l cccc  ccc  cccc | cccc  ccc  cc}
\toprule

\multirow{3}{*}{Method} & \multicolumn{11}{c|}{InsAVE-80K} & \multicolumn{9}{c}{AvED-Bench} \\
\cmidrule(lr){2-12} \cmidrule(lr){13-21}
& \multicolumn{4}{c|}{Video Eval} & \multicolumn{3}{c|}{Audio Eval} & \multicolumn{4}{c|}{A-V Eval} & \multicolumn{4}{c|}{Video Eval} & \multicolumn{3}{c|}{Audio Eval} & \multicolumn{2}{c}{A-V Eval} \\
& FVD$\downarrow$ & TV-A$\uparrow$ & TC$\uparrow$ & SSIM$\uparrow$ & FAD$\downarrow$ & TA-A$\uparrow$ & LPAPS$\downarrow$ & AV-A$\uparrow$ & PEAVS$\uparrow$ & S-C$\uparrow$ & S-D$\downarrow$ & FVD$\downarrow$ & TV-A$\uparrow$ & TC$\uparrow$ & SSIM$\uparrow$ & FAD$\downarrow$ & TA-A$\uparrow$ & LPAPS$\downarrow$ & AV-A$\uparrow$ & PEAVS$\uparrow$ \\
\midrule
\multicolumn{21}{c}{Comparison with State-of-the-Art Methods} \cr
\midrule
AvED & 305.41 & 24.68 & 94.73 & 90.78 & 2.86 & 29.89 & 1.80 & 26.44 & 3.15 & 2.98 & 9.42 & 538.99 & 24.08 & 94.01 & 73.09 & 4.68 & 31.50 & 1.72 & 23.40 & 2.94 \\
CoherentAVEdit & 326.63 & 25.18 & 95.60 & 93.26 & 8.01 & 32.47 & 2.13 & 22.67 & 3.20 & 1.25 & 12.55 & 529.85 & 24.13 & 95.18 & 91.96 & 7.72 & 34.19 & 1.93 & 20.86 & 2.80 \\
AVI-Edit & 205.34 & 25.15 & 95.45 & 90.66 & 5.09 & 30.07 & 1.87 & 26.37 & 2.54 & 4.06 & 9.25 & 372.37 & 23.79 & 95.24 & 89.20 & 7.65 & 29.35 & 1.67 & 23.21 & 2.50 \\
InstructAV2AV (Ours) & \textbf{180.38} & \textbf{25.23} & \textbf{95.65} & \textbf{93.84} & \textbf{2.75} & \textbf{34.96} & \textbf{1.77} & \textbf{27.72} & \textbf{3.24} & \textbf{4.67} & \textbf{8.96} & \textbf{227.82} & \textbf{24.26} & \textbf{95.27} & \textbf{92.72} & \textbf{4.32} & \textbf{35.89} & \textbf{1.58} & \textbf{23.71} & \textbf{3.20} \\ \midrule
\multicolumn{21}{c}{Ablation Study} \cr
\midrule
\textit{W/o} SC & 467.20 & 25.08 & 95.58 & 49.97 & 4.20 & 34.00 & 1.96 & 26.57 & 3.21 & 4.56 & 9.48 & 654.16 & 23.87 & 94.10 & 39.90 & 7.71 & 32.50 & 1.79 & 22.28 & 3.15 \\
\textit{W/o} SIGA & 187.28 & 24.90 & 95.52 & 93.65 & 3.26 & 34.50 & 1.88 & 26.40 & 3.05 & 4.38 & 9.16 & 314.82 & 23.10 & 95.15 & 86.29 & 6.58 & 34.17 & 1.65 & 20.99 & 2.70 \\
\textit{W/o} TSTS & 291.55 & 24.87 & 94.96 & 85.63 & 5.18 & 34.80 & 2.05 & 25.12 & 3.19 & 4.12 & 9.56 & 327.97 & 23.40 & 94.51 & 72.78 & 6.37 & 33.51 & 1.84 & 20.82 & 3.11 \\
\bottomrule
\end{tabular}
\end{adjustbox}
\end{center}
\end{table*}

\begin{table}[t] 
\caption{Percentage (\%) of user preference study results.}
  \vspace{-3pt}
\label{tab:compare-user-study}
\centering 
\footnotesize
\setlength\tabcolsep{6pt} 
\begin{adjustbox}{width=0.48\textwidth,keepaspectratio}
\begin{tabular}{l | ccc | ccc} 
\toprule 
\multirow{2}{*}{Method} & \multicolumn{3}{c|}{\sc InsAVE-80K} & \multicolumn{3}{c}{AvED-Bench} \\
& AVS & TA & OP & AVS & TA & OP \\ 
\midrule
AvED & 1.40 & 2.60 & 0.80 & 1.80 & 3.40 & 2.20 \\
CoherentAVEdit & 17.00 & 20.60 & 18.60 & 18.40 & 23.20 & 23.00 \\
AVI-Edit & 32.60 & 31.40 & 34.00 & 33.00 & 31.60 & 31.20 \\
InstructAV2AV (Ours) & \textbf{49.00} & \textbf{45.40} & \textbf{46.60} & \textbf{46.80} & \textbf{41.80} & \textbf{43.60} \\
\bottomrule
\end{tabular}
\end{adjustbox}
\end{table}

\subsection{Experimental Setups}

\paragraph{Training details.}
We initialize our model from the pre-trained weights of the audio-video generation model~\cite{ovi} and train it at a resolution of 720p for audio-visual editing. Following our two-stage strategy, the modality-specific training stage independently optimizes the video and audio branches for 320K and 960K steps, respectively. Subsequently, the joint audio-video training stage finetunes the unified architecture for an additional 400K steps. All experiments are conducted on 8 NVIDIA H100 GPUs, using the AdamW~\cite{adamw} optimizer with a learning rate of $1\times10^{-5}$.

\paragraph{Evaluation datasets.}
In addition to the 1K human-curated evaluation split from our proposed InsAVE-80K, we assess the zero-shot generalization of our framework on publicly available samples from AvED-Bench~\cite{aved}.

\subsection{Comparison with State-of-the-Art Methods}
Instruction-guided audio-video joint editing is a highly emerging field. To the best of our knowledge, we conduct a comparison against all currently available methods in this domain: AvED~\cite{aved}, AVI-Edit~\cite{aviedit}, and CoherentAVEdit~\cite{coherentavedit}. All methods are evaluated under their default configurations. 
Notably, as AVI-Edit and CoherentAVEdit additionally rely on instance masks, we provide them with ground-truth annotations, granting them a substantial privilege during evaluation.

\paragraph{Quantitative comparisons.}
We conduct quantitative comparisons from three complementary aspects: video, audio, and joint audio-visual performance.
\textit{(i)} For the video modality, we measure generation quality via Fréchet Video Distance (FVD)~\cite{fvd}, text-video alignment (TV-A) using VideoCLIP-XL~\cite{videoclipxl}, temporal consistency (TC) through consecutive-frame CLIP similarity~\cite{clip}, and background preservation with the Structural Similarity Index Measure (SSIM)~\cite{ssim}.
\textit{(ii)} For the audio modality, we assess fidelity via Fréchet Audio Distance (FAD)~\cite{fad}, text-audio alignment (TA-A) with CLAP~\cite{clap}, and acoustic consistency using LPAPS~\cite{lpaps}.
\textit{(iii)} For the joint audio-visual domain, we examine cross-modal alignment (AV-A) utilizing ImageBind~\cite{imagebind}, overall synchronization quality via PEAVS~\cite{peavs}, and lip synchronization measured by Sync-C (S-C) and Sync-D (S-D)~\cite{sync}.
As shown in Table~\ref{tab:quantitative_combined}, our model outperforms relevant methods across all metrics. We provide the detailed definitions of evaluation protocols in the supplementary material.

\paragraph{Qualitative comparisons.}
As shown in Figure~\ref{fig:compare}, InstructAV2AV demonstrates significant visual and acoustic improvements over state-of-the-art methods~\cite{aviedit, aved, coherentavedit}.
Specifically, AvED~\cite{aved} struggles to maintain visual temporal consistency and audio fidelity, leading to severe frame-wise flickering and disturbing audio artifacts (Figure~\ref{fig:compare}, left and right).
CoherentAVEdit~\cite{coherentavedit} fails to synthesize the target speech (Figure~\ref{fig:compare}, left), while AVI-Edit struggles with preserving the non-target flame sound (\eg, Figure~\ref{fig:compare}, right).
In contrast, our InstructAV2AV accurately modifies the target audio-visual content while faithfully preserving non-target contexts.

\paragraph{User preference study.}
To further evaluate perceptual quality, we conduct a user study assessing human preference. Given the source audio-video inputs and editing instructions, participants are presented with the edited results from InstructAV2AV and the comparison methods, and are asked to choose the best result based on three criteria:
\textit{(i)} \textbf{Audio-Video Synchronization (AVS):} The output with the most precise temporal and semantic alignment across modalities.
\textit{(ii)} \textbf{Text Alignment (TA):} The output that most accurately follows the provided editing instruction.
\textit{(iii)} \textbf{Overall Preference (OP):} The most preferred clip considering holistic audio-visual quality.
We randomly select 20 samples from each evaluation dataset and recruit 25 volunteers to provide independent evaluations. As shown in Table~\ref{tab:compare-user-study}, our model significantly outperforms the baselines, achieving the highest preference scores across all evaluated dimensions.

\begin{figure*}[t]
  \centering
  \includegraphics[width=\linewidth]{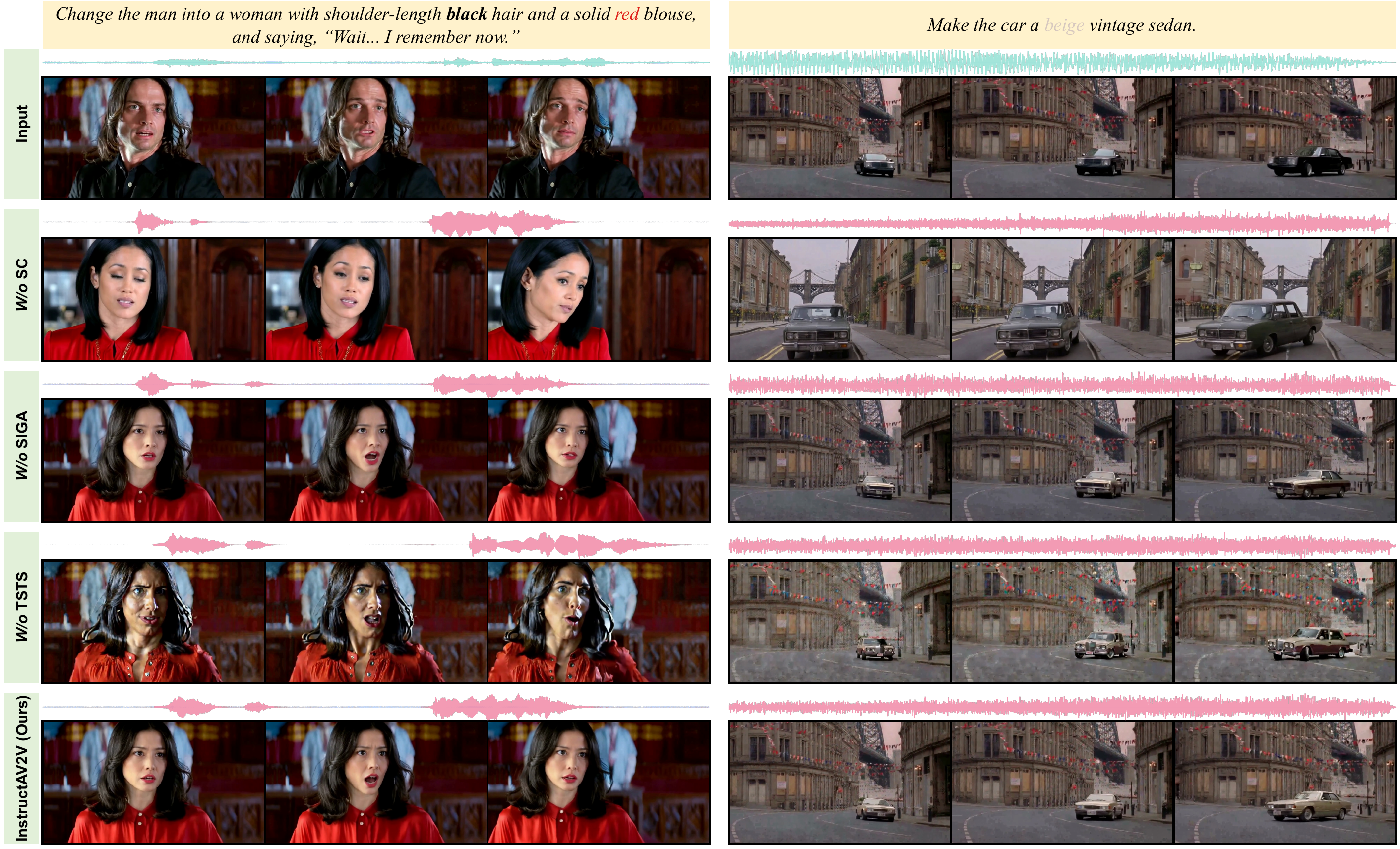}
  \vspace{-6mm}
  \caption{
   Ablation study results of different InstructAV2AV variants.
  }
  \label{fig:ablation}
  \vspace{-2mm}
\end{figure*}

\subsection{Ablation Study}
To evaluate the effectiveness of key components of InstructAV2AV, we conduct ablation studies as shown in Table~\ref{tab:quantitative_combined} and Figure~\ref{fig:ablation}.

\paragraph{W/o Source Concatenation (SC).}
We remove the concatenation of the source latent codes and feed the noisy latent codes directly into the framework. Without this structural source condition, the framework fails to preserve non-target contexts (\eg, the drastically altered backgrounds in Figure~\ref{fig:ablation}, left and right).

\paragraph{W/o Source-Instruction Gated Attention (SIGA).}
We replace the proposed SIGA module with standard text cross-attention. This significantly degrades instruction-following capabilities, leading to audio hallucinations and stuttering artifacts (\eg, an unnatural extra ``t'' sound after the word ``wait'' in Figure~\ref{fig:ablation}, left).

\paragraph{W/o Two-Stage Training Strategy (TSTS).}
We discard the two-stage training strategy and directly fine-tune the entire framework. This abrupt optimization exposes the model to a large domain gap, yielding visually distorted and inconsistent editing results (\eg, the severe abnormal lighting in Figure~\ref{fig:ablation}, left and right).

\begin{figure*}[t]
  \centering
  \includegraphics[width=\linewidth]{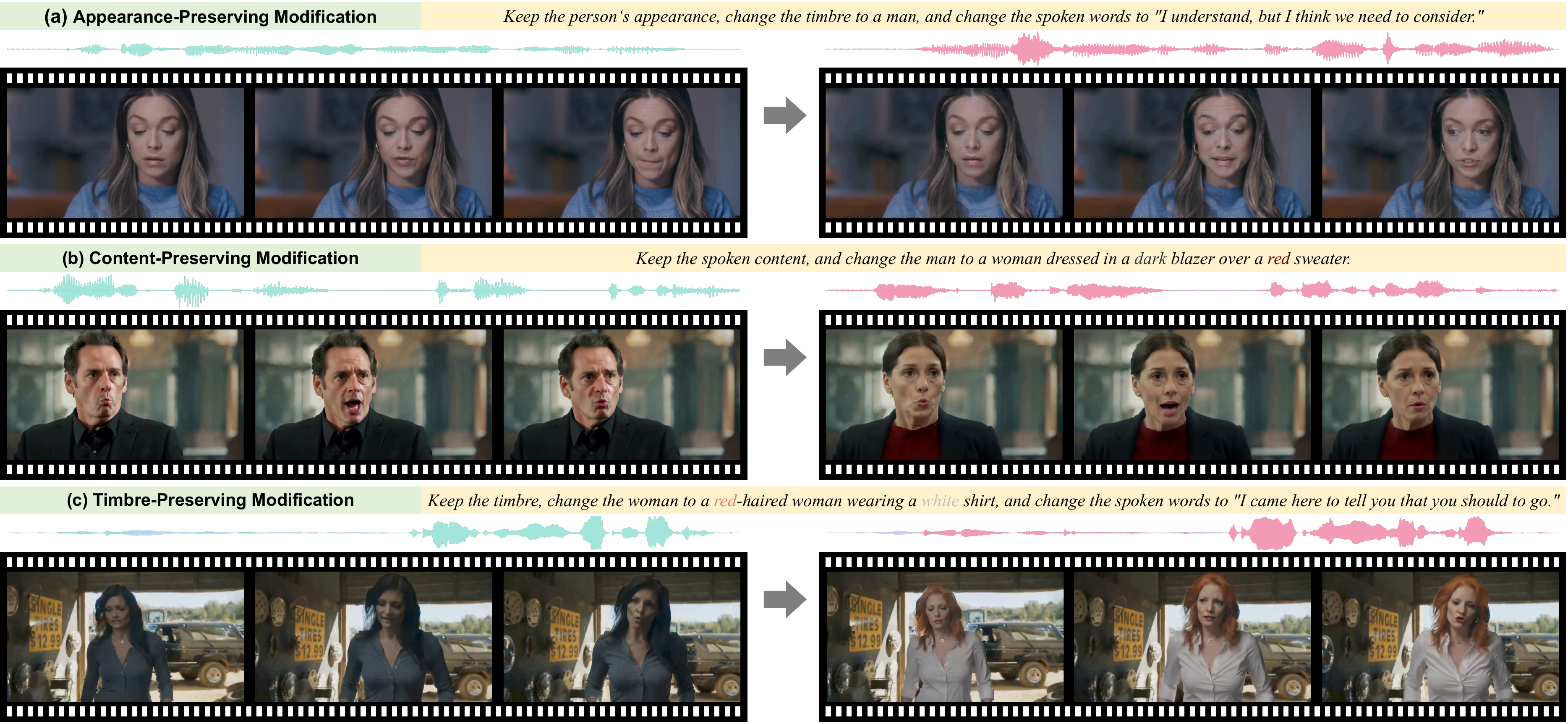}
  \caption{Examples of additional application scenarios for our proposed InstructAV2AV framework.}
  \label{fig:more1}
\end{figure*}

\subsection{Applications}
In addition to the four application scenarios illustrated in Figure~\ref{fig:teaser} (\ie, identity-preserving speech modification, audio-visual instance editing, audio-visual instance insertion, and audio-visual instance removal), we present three additional applications:
\textit{(i)} Appearance-preserving modification: preserving the visual appearance of the target speaker while jointly modifying the spoken content and vocal timbre (Figure~\ref{fig:more1}~(a));
\textit{(ii)} Content-preserving modification: preserving the original spoken content while jointly modifying the visual appearance and vocal timbre (Figure~\ref{fig:more1}~(b)); and
\textit{(iii)} Timbre-preserving modification: preserving the target vocal timbre while jointly modifying the spoken content and speaker appearance (Figure~\ref{fig:more1}~(c)).
By independently manipulating these fine-grained attributes, we demonstrate the framework's robust controllability, precise editing capabilities, and strong attribute disentanglement.

\begin{figure*}[t]
  \centering
  
  \includegraphics[width=\linewidth]{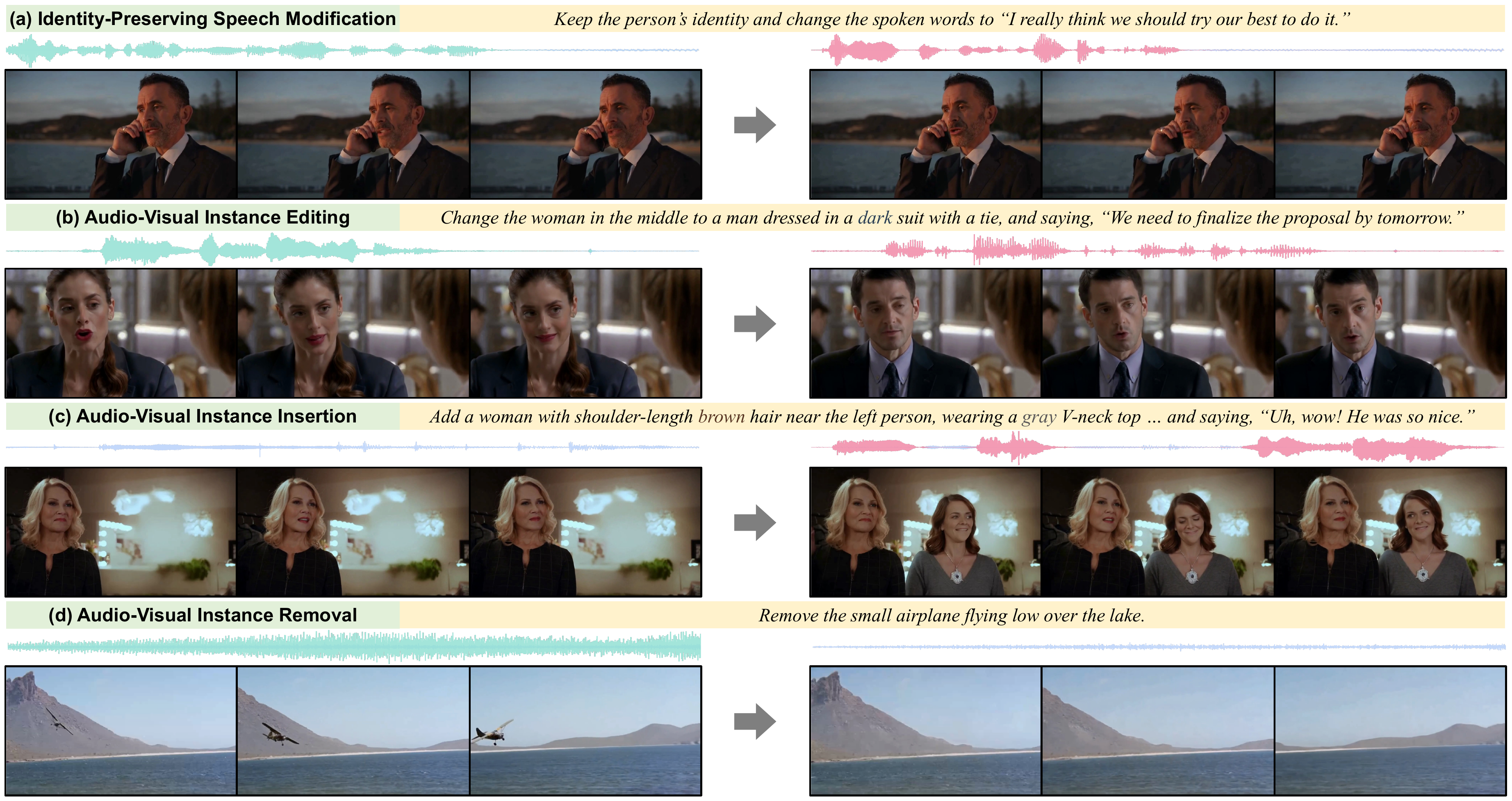}
  \vspace{-6mm}
  \caption{Additional qualitative results of InstructAV2AV.}
  \label{fig:more2}
\end{figure*}

\section{Conclusion and Limitation}
In this paper, we propose InstructAV2AV, the first end-to-end framework for open-world audio-video joint editing, using solely text instructions.
To provide sufficient high-quality training data, we design a scalable data synthesis pipeline and construct InsAVE-80K, the first large-scale audio-video editing dataset with source-to-target pairs. 
Architecturally, we build our framework upon a foundational audio-video generation model. To anchor the source context, we concatenate the audio-video source with noisy latent codes for each modality. We further introduce the source-instruction gated attention module to improve content preservation and instruction following. Additionally, a tailored two-stage training strategy is proposed to achieve smooth model convergence. 
Extensive experiments demonstrate that InstructAV2AV consistently outperforms state-of-the-art baselines across 11 metrics spanning three aspects on two evaluation sets. By enabling fine-grained attribute disentanglement and flexible control, we believe our framework paves the way for highly controllable audio-visual content creation.

As our framework is built upon a foundational audio-video generation model, the quality of the edited results is inherently tied to the underlying backbone, thus inheriting its common failure modes. For instance, the model may struggle with maintaining physical realism during complex object interactions, ensuring lighting consistency in multi-source environments, and preserving 3D spatial consistency or object permanence during extensive camera movements. However, we believe that with the rapid advancement of the generative AI field, these base model limitations will be progressively alleviated. Furthermore, our framework is highly flexible and can be seamlessly integrated with future state-of-the-art generation models to directly benefit from their improvements.

\section{Appendix}

\subsection{Editing Capability Comparison}
\label{sec:model_comparison}

We present a capability comparison between InstructAV2AV and existing methods in Table~\ref{tab:model_comparison}. To the best of our knowledge, InstructAV2AV is the first end-to-end open-world audio-video joint editing framework using solely text instructions. Specifically, our framework demonstrates the following two key advantages:

\paragraph{Mask-free editing.} Compared to CoherentAVEdit \cite{coherentavedit}, Object-AVEdit~\cite{object_avedit}, and AVI-Edit~\cite{aviedit}, which rely on explicit instance masks or bounding boxes for spatial grounding, InstructAV2AV operates solely on text instructions to achieve precise instruction following and content preservation for audio-video joint editing.

\paragraph{End-to-end editing.} Compared to One-Shot~\cite{liang2024language}, AvED~\cite{aved}, CoherentAVEdit~\cite{coherentavedit}, and Object-AVEdit~\cite{object_avedit}, which lack end-to-end training, InstructAV2AV achieves strict temporal and semantic synchronization within a unified end-to-end framework, effectively handling diverse open-world scenarios (\eg, human speech and general audio).

\begin{table*}[t]
\caption{Capability comparison of audio-video editing methods.}
\label{tab:model_comparison}
\centering
\small
\setlength\tabcolsep{4pt} 
\begin{tabular}{l ccccccc}
\toprule
\textbf{Method} & Train-based & End-to-end & Mask-free & Text-guided  & Human speech & General audio \\
\midrule
One-Shot~\cite{liang2024language} & \texttimes & \texttimes & \checkmark & \checkmark  & \texttimes & \checkmark \\
AvED~\cite{aved} & \texttimes & \texttimes & \checkmark & \checkmark &  \texttimes & \checkmark \\
CoherentAVEdit~\cite{coherentavedit} & \checkmark & \texttimes & \texttimes & \checkmark  & \texttimes & \checkmark \\
Object-AVEdit~\cite{object_avedit} & \checkmark & \texttimes & \texttimes & \checkmark  & \texttimes & \checkmark \\
AVI-Edit~\cite{aviedit} & \checkmark & \checkmark & \texttimes & \checkmark &  \checkmark & \checkmark \\
\midrule
\textbf{InstructAV2AV (Ours)} & \checkmark & \checkmark & \checkmark & \checkmark  & \checkmark & \checkmark \\
\bottomrule
\end{tabular}
\end{table*}

\subsection{Detailed Metric Definitions}
\label{sec:metric_details}

We provide complete definitions and implementation details for all evaluation metrics used in our experiments, organized into three complementary aspects: video modality, audio modality, and joint audio-video domain.

\subsubsection{Video Metrics}

\paragraph{Fr\'{e}chet Video Distance (FVD).}
We adopt the FVD metric~\cite{fvd} to evaluate the overall visual realism and spatiotemporal coherence of the generated results. Using a VideoMAE~\cite{videomae} ViT-g/Hybrid backbone, we extract features from all frames in each video at a resolution of $256 \times 256$ with a temporal stride of 1. The FVD score is then computed as the Fr\'{e}chet distance between the real and generated feature distributions:
\begin{equation}
    \mathrm{FVD} = \left\| \mu_\mathrm{r} - \mu_\mathrm{g} \right\|_2^2 + \mathrm{Tr}\left(\Sigma_\mathrm{r} + \Sigma_\mathrm{g} - 2\left(\Sigma_\mathrm{r} \Sigma_\mathrm{g}\right)^{1/2}\right),
\end{equation}
where $\mu$ and $\Sigma$ denote the mean and covariance of the VideoMAE feature distributions computed over all real and generated samples, respectively.

\medskip

\paragraph{Text-Video Alignment (TV-A).}
TV-A quantifies the semantic alignment between the editing instruction and the edited video using VideoCLIP-XL~\cite{videoclipxl}. For each video, we uniformly sample 8 frames, resize them to $224 \times 224$, and extract visual features via the VideoCLIP-XL vision encoder. The text instruction is encoded by the VideoCLIP-XL text encoder. Both feature vectors are $\ell_2$-normalized, and the alignment score is their cosine similarity:
\begin{equation}
    \mathrm{TV\text{-}A} = \cos\left(e_\mathrm{text},\, e_\mathrm{video}\right).
\end{equation}

\medskip

\paragraph{Temporal Consistency (TC).}
TC measures the smoothness and coherence of the edited video across frames. We encode every frame of the edited video using CLIP~\cite{clip} ViT-B/32 and $\ell_2$-normalize the resulting image features. We report the mean cosine similarity between temporally adjacent frames:
\begin{equation}
    \mathrm{TC} = \frac{1}{T-1}\sum_{t=1}^{T-1} \cos\left(e_\mathrm{frame}^{(t)},\, e_\mathrm{frame}^{(t+1)}\right).
\end{equation}

\medskip

\paragraph{Structural Similarity (SSIM).}
We compute the SSIM between the source and edited video frames to evaluate the overall structural accuracy. For each frame $t$, we compute the full-frame SSIM score using the \texttt{skimage} library, and then average the SSIM values across all frames:
\begin{equation}
    \mathrm{SSIM} = \frac{1}{T}\sum_{t=1}^{T} \mathrm{SSIM_{score}}(t).
\end{equation}

\medskip

\subsubsection{Audio Metrics}

\paragraph{Fr\'{e}chet Audio Distance (FAD).}
We compute FAD using the official \texttt{frechet\_audio\_distance} library with the VGGish backbone~\cite{vggish}. Audio waveforms are resampled to 16\,kHz. We disable PCA and activation clipping to obtain raw VGGish embeddings. The FAD score is the Fr\'{e}chet distance between the real and generated audio feature distributions, following the same formulation as FVD.

\medskip

\paragraph{Text-Audio Alignment (TA-A).}
TA-A measures the semantic consistency between the editing instruction and the edited audio using CLAP~\cite{clap}. We encode the edited audio waveform and the text instruction via the CLAP audio and text encoders, respectively. Both embeddings are $\ell_2$-normalized, and the alignment score is their cosine similarity:
\begin{equation}
    \mathrm{TA\text{-}A} = \cos\left(e_\mathrm{text},\, e_\mathrm{audio}\right).
\end{equation}

\medskip

\begin{figure*}[t]
    \centering
    \includegraphics[width=\linewidth]{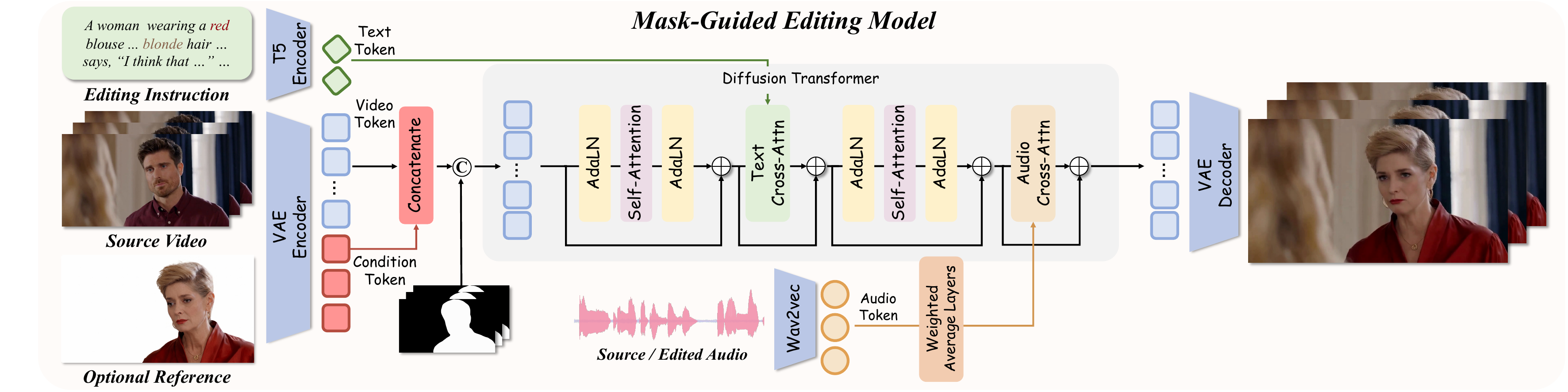}
    \caption{Illustration of our mask-guided audio-sync video editing data engine.
    Three conditioning mechanisms are integrated into the Wan2.2-5B backbone: mask concatenation for spatial grounding, EMO2-style audio cross-attention for temporal synchronization, and reference token concatenation for appearance preservation.}
    \label{fig:data_engine}
\end{figure*}

\paragraph{Learned Perceptual Audio Patch Similarity (LPAPS).} 
In addition to the CLAP-level similarity, we compute LPAPS~\cite{lpaps} as a layer-wise perceptual distance. We extract hidden states from all layers of the CLAP audio encoder for both source and edited audio, $\ell_2$-normalize each hidden state, and compute the spatially averaged $\ell_2$ distance across layers:
\begin{equation}
    \mathrm{LPAPS} = \sum_{l=1}^{L} \mathrm{Avg}\left( \left\| \hat{h}_l^\mathrm{src} - \hat{h}_l^\mathrm{edit} \right\|_2^2 \right),
\end{equation}
where $\hat{h}_l$ denotes the $\ell_2$-normalized hidden state at layer $l$, and $\mathrm{Avg}(\cdot)$ denotes averaging over all spatial dimensions.

\subsubsection{Audio-Video Metrics}

\paragraph{Audio-Video Alignment (AV-A).}
AV-A evaluates the semantic coherence between the edited video and edited audio using ImageBind~\cite{imagebind}. We encode the edited video and audio using ImageBind-huge to obtain modality-specific embeddings. Both embeddings are $\ell_2$-normalized, and the alignment score is the cosine similarity between the video and audio embeddings:
\begin{equation}
    \mathrm{AV\text{-}A} = \cos\left(e_\mathrm{video}^\mathrm{IB},\, e_\mathrm{audio}^\mathrm{IB}\right).
\end{equation}

\medskip

\paragraph{Perceptual Evaluation of Audio-Visual Synchrony (PEAVS).}
PEAVS measures the perceptual quality of audio-video synchronization using the official \texttt{avgen-eval-toolkit}~\cite{peavs}. We run the PEAVS feature extraction pipeline followed by the metric prediction script, which outputs a predicted mean opinion score for audio-visual synchronization quality.

\medskip

\paragraph{Sync-C (S-C) and Sync-D (S-D).}
Sync-C and Sync-D are lip synchronization metrics derived from SyncNet~\cite{sync}. We use the SyncNet pipeline with a batch size of 20, a temporal search radius of \texttt{vshift=15} frames, face detection scale of 0.25, and crop scale of 0.40. Sync-C measures the confidence of lip-audio correspondence (higher is better), while Sync-D quantifies the desynchronization distance in milliseconds (lower is better).

\subsection{Data Engine: Mask-Guided Editing Model}
\label{sec:data_engine_details}
In this section, we provide a detailed description of the mask-guided audio-sync video instance editing model used in our data synthesis pipeline (Sec~\ref{subsec:data_engine_1}).

\subsubsection{Model Architecture}~\\
Our data engine builds upon Wan2.2-5B~\cite{wan2.2}, a pre-trained video diffusion model. To enable audio-sync mask-guided editing, we introduce three modifications, as shown in Figure~\ref{fig:data_engine}.

\medskip

\paragraph{Mask-guided spatial conditioning.}
Given a source video clip and its corresponding entity mask $M \in \{0,1\}^{T \times H \times W}$ produced by Grounded-SAM-2~\cite{grounded_sam2}, we inject the mask into the denoising process to explicitly constrain spatial modifications. Specifically, the mask is concatenated with the noisy latent codes along the channel dimension before being fed into the diffusion transformer:
\begin{equation}
    \hat{z}_t = \mathrm{Concat}\left(z_t,; \mathrm{Downsample}(M)\right),
\end{equation}
where $\mathrm{Downsample}(\cdot)$ maps the binary mask to the spatial resolution of the latent code.

\medskip

\paragraph{Audio feature injection.}
To ensure temporal and semantic synchronization between the edited video and audio, we incorporate audio features into the video backbone via frame-wise cross-attention. Following EMO2~\cite{emo2}, we apply a layer-wise weighted average to aggregate features from all layers of the audio encoder:
\begin{equation}
    \bar{e}_\mathrm{audio}^{(t^\text{th frame})} = \sum_{l=1}^{L} \alpha_l \cdot e_\mathrm{audio}^{(l, t^\text{th frame} )},
\end{equation}
where $\alpha_l$ are learnable scalar weights, and $e_\mathrm{audio}^{(l, t\text{th frame})}$ is the audio feature corresponding to the $t$-th video frame at layer $l$. The aggregated audio representation is then injected into each video transformer block through frame-wise cross-attention, enabling each video frame to attend to its corresponding audio context.

\medskip

\paragraph{Optional appearance reference.}
To support appearance-preserving editing scenarios (\eg, changing only the motion or sound while keeping visual identity consistent), we introduce an optional reference conditioning mechanism. Given a reference image $I_\mathrm{ref}$ that specifies the desired appearance, we encode it into a sequence of visual tokens $f_\mathrm{ref} \in \mathbb{R}^{N_\mathrm{ref} \times D}$ using the video VAE encoder. These reference tokens are then concatenated with the noisy video tokens before being fed into the diffusion transformer:
\begin{equation}
\hat{f}_t = \left[f_\mathrm{ref}^{(1)}; \ldots; f_\mathrm{ref}^{(N_\mathrm{ref})}; f_t^{(1)}; \ldots; f_t^{(T \cdot H' \cdot W')}\right],
\end{equation}
where $f_t$ are the noisy video tokens at timestep $t$. 
During training, the reference image is sampled from the target video with a configurable dropout rate to maintain diversity. At inference, this mechanism is optional: when no reference is provided, the model degenerates to standard instruction-guided editing.

\subsubsection{Training Strategy}~\\
Following AVI-Edit~\cite{aviedit}, we train the mask-guided editing model using a flow matching objective. 
We train the model to reconstruct the source video conditioned on the source audio, instance mask, reconstruction instruction, and unmasked background regions.
Consequently, during inference, the model can generate edited content when conditioned on the target edited audio, a random mask, and diverse editing instructions.

Given the source video latent $z_1^{\mathrm{v}}$, source audio $X$, entity mask $M$, reconstruction instruction $I$, and a noise sample $\epsilon \sim \mathcal{N}(0, \mathbf{I})$, we construct the noisy latent as:
\begin{equation}
    z_t = (1-t)\epsilon + t z_1^{\mathrm{v}}, \quad t \in [0,1].
\end{equation}
The training objective is:
\begin{equation}
    \mathcal{L}_\mathrm{FM} = \mathbb{E}_{t, \epsilon, z_1^{\mathrm{v}}} \left[ \left\| v_\theta\left(t, z_t, M, X, I\right) - (z_1^{\mathrm{v}} - \epsilon) \right\|_2^2 \right].
\end{equation}
This training strategy requires no manually annotated edited video targets, enabling scalable data generation for our InstructAV2AV.

\subsection{Organization of Supplementary Video}
We provide a supplementary video to dynamically showcase our instruction-guided audio-video joint editing results. The video is structured as follows:
\textit{(i)} \textbf{Representative application scenarios:} We demonstrate four representative editing scenarios to highlight the diverse and compelling applications of our framework (Figure~\ref{fig:teaser} and Figure~\ref{fig:more2}).
\textit{(ii)} \textbf{Additional application scenarios:} We showcase three extended application scenarios to further demonstrate the versatility of our framework (Figure~\ref{fig:more1}).
\textit{(iii)} \textbf{Comparison with state-of-the-art methods:} We provide audio-video joint editing comparisons against state-of-the-art methods (Figure~\ref{fig:compare}).
\textit{(iv)} \textbf{Ablation study:} Finally, we present ablation results to validate the effectiveness of our core components (Figure~\ref{fig:ablation}).

\bibliographystyle{ACM-Reference-Format}
\bibliography{bibliography}

\clearpage

\end{document}